\documentclass[10pt,conference,a4paper]{IEEEtran}
\usepackage{times,amsmath,epsfig}
\newcommand{\degree}{\ensuremath{^\circ}}
\title{Computer Aided ECG Analysis - State of the Art and Upcoming Challenges}
\author{%
{Marko Velic{\small $~^{1}$}, Ivan Padavic{\small $~^{2}$}, Sinisa Car{\small $~^{3}$} }%
\vspace{1.6mm}\\
\fontsize{10}{10}\selectfont\itshape
$^{1}$\,University Computing Centre, University of Zagreb\\
Josipa Marohnica 5, 10000 Zagreb Croatia\\
\fontsize{9}{9}\selectfont\ttfamily\upshape
marko@velic.biz
\vspace{1.2mm}\\
\fontsize{10}{10}\selectfont\rmfamily\itshape
$^{2}$\,Trikoder Ltd.\\
Draskoviceva 80, 10000 Zagreb Croatia\\
\fontsize{9}{9}\selectfont\ttfamily\upshape
ivan.padavic@trikoder.net
\vspace{1.2mm}\\
\fontsize{10}{10}\selectfont\rmfamily\itshape
$^{3}$\,General Hospital Varazdin, Cardiology Department\\
Mestroviceva b.b., 42000 Varazdin Croatia\\
\fontsize{9}{9}\selectfont\ttfamily\upshape
sinisa.car@vz.t-com.hr
}
\begin{document}
\maketitle
\begin{abstract}
In this paper we present current achievements in computer aided ECG analysis and their applicability in real world medical diagnosis process. Most of the current work is covering problems of removing  noise, detecting heartbeats and rhythm-based analysis. There are some advancements in particular ECG segments detection and beat classifications but with limited evaluations and without clinical approvals. This paper presents state of the art advancements in those areas till present day. Besides this short computer science and signal processing literature review, paper covers future challenges regarding the ECG signal morphology analysis deriving from the medical literature review. Paper is concluded with identified gaps in current advancements and testing, upcoming challenges for future research and a bullseye test is suggested for morphology analysis evaluation.
\end{abstract}

\section{Introduction}
In computer aided ECG signal analysis, many of the ECG interpretation problems are solved but there are still doubts if those methods are really useful for practical purposes. Big advancement were made in noise removal, heartbeat detection (QRS detectors), heart rate variability (HRV) analysis and recently even some classification of the ECG shapes. As algorithms are becoming more and more powerful and precise, gaps between recent algorithmic advancements and available testing methodology are becoming to emerge. This paper is presenting short analysis of the current state of the art. Comprehensive literature review is not goal of this paper and a reader is advised to follow given references on recent work for deeper information on each mentioned subject. In this paper more attention is put on the medical literature where new challenges are identified going beyond the current advancements. Due to that, gaps in the testing methodology are identified and new approaches are suggested. In respect with size constraints, this paper is not showing figures for every mentioned ECG issue and reader is here again advised to follow page-level references on medical literature.
Goal of this paper is not to serve as a cookbook for ECG waves interpretation nor as a recipe for ECG algorithm heuristic development but to indicate clinical importance of the ECG morphology recognition problem and pave the way for future research through concrete and medically plausible cases.

\section{Medical Importance of ECG Analysis}

Electrocardiogram (ECG or EKG) is a record of bio-electric potential variation recorded through time on the body surface that represents heart beats \cite {EXSY:EXSY444}. Every heartbeat cycle is normally characterized by the sequence of waveforms known as a P wave, QRS complex and a T wave. Time intervals between those waveforms as well as their shapes and orientation are representing physiological processes occurring in heart and autonomous nervous system. Although today in medical centres advanced equipment and tools are used for detecting heart-beat arrhythmias and other cardiovascular abnormalities, visual inspection of the multi-channel (lead) ECG record is still the first step taken by cardiologists in diagnosis process \cite{Ghaffari:2008}.

\begin{figure}[h]
	\centerline{\psfig{figure=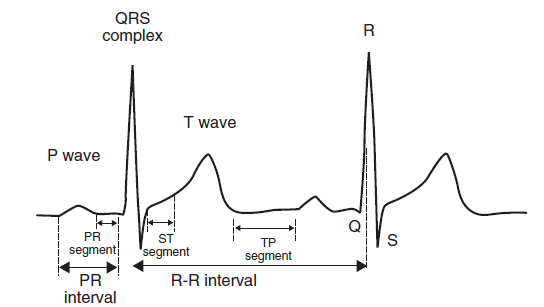,width=68.7mm} }
	\caption{Normal ECG signal with marked characteristics, according to \cite{Kabir2012481:2012}}
	\label{fig:sample_graph}
\end{figure}

Detailed explanation of the physiological process behind the ECG signal shape is out of the scope of this paper, but for the easier understanding of the main goals of this research we will give short explanation and a reader is advised to follow given references for detailed information. 

Human heart is divided into four main chambers called atria and ventricles both with their left and right instances. Those chambers together form a biological pump for propelling the blood throughout the body. Besides those four obvious sections there are some other parts of the heart for very specialized functions like dividing atria from ventricles, slow impulse propagation, very fast impulse propagation etc., all of them performing particular tasks, ensuring that blood flows properly and efficiently throughout the body. When electrical impulse propagates through heart and all these specialized cells, ECG electrodes pick up that impulse in various directions and speed. In this way ECG waveforms are formed \cite{Clifford:2006}, \cite{camm2009esc}. With that in mind one can logically assume that different problems in different kind of cells or different parts of heart will have corresponding effects in ECG waves direction and morphology \cite {de2008basic} and this connection will be covered in the following chapters.

Efficient and fast ECG analysis algorithms are needed in clinical practice but also in pre-hospital use cases since clinical findings indicated that there was a significant improvement in patient outcome based on this early treatment \cite{Purvis1999604}. Pre-hospital ECG is a test that may potentially influence the management of patients with acute myocardial infarction through wider, faster in-hospital utilization of re-perfusion strategies and greater usage of invasive procedures, factors that may possibly reduce short term mortality \cite{Canto1997498}. 
Medical literature suggests clinical importance of ECG not only in identifying heart problems itself, but also other health issues that leave a trace on ECG as a symptomatic phenomena like ECG patterns reflecting antidepressant treatment \cite{doi:10.3109/08039489309104126}.

\section{State of the Art}

\subsection{Noise Removal}

As is the case with any other signal, pre-processing of the ECG data is an important step in the analysis \cite{Ghosh_Raychaudhuri_2007}. Filtering of the data is important because of the noise that in the case of ECG recordings can have several causes like interference with other devices and signals, infrastructure noise, muscle movements under the skin where electrodes are placed, respiratory movements or even friction between the skin and the electrodes. Since the nature of the most of the mentioned noise sources is known and their features like frequency components and periodicity are understood those features can be used for noise reduction or even complete attenuation. Due to that, we have today a wide spectrum of noise removal techniques like band pass filters \cite{manikandan2012novel}, Fourier based analysis and transforms \cite{Darrington2009}, wavelet transformations \cite{mishra2010real}, cubic splines \cite{kabir2012denoising} etc. These methods rely on techniques like averaging and smoothing of the signal or transforming the signal from time-domain to frequency-domain and then removing the noise frequency components. Results of these techniques are reported in the literature and they show that noise removal in ECG signal analysis is well coped.

\subsection{QRS Detection}

Biomedical signal processing and pattern recognition literature offers a wide spectrum of solutions and approaches to QRS detection. Even as early as in 60s of the 20th century, algorithms like AZTEC  \cite{yanowitz1974accuracy} and Pan-Tompkins resulted in very high QRS detection rates. 

Basic principle of the AZTEC algorithm was inclination measurement i.e. amplitude analysis. Decision about the arrhythmia occurrence was made based on the intervals between detected heart beats. At that time, there was no standardised ECG testing database and algorithm evaluation was performed in the long-term study on six patients and short-term study on thirty patients. QRS detection rate was 99\%. Interesting result is that this algorithm was capable of detecting Premature Ventricular Beats (PVB) with 90\% detection rate and 1\% of false positive PVBs. Nevertheless it is important to mention that AZTEC had one exception built in - when major noise was detected, classification was not performed \cite{Aztec_4502549}. 

Pan Tompkins algorithm uses several mathematical transforms to find out width, amplitude and inclination of the particular ECG signal artefacts. This algorithm works in three phases - implementation of the linear digital filter, squaring of the signal to ensure that all signal segments are in the set of positive numbers and finally thresholds and logic for QRS segment detection. Algorithm needs a learning phase and overall accuracy measured on the MIT-BIH Arrhythmia Database \cite{mark1988bih} was 99.3\%. \cite{pan1985real}.

In recent years we can find fast and robust modern approaches based on various classifiers like Neural Networks \cite{NN_QRS_921381} or Support Vector Machines \cite{Mehta_2008}. These approaches are trying to pre process the signal to remove the noise and find distinctive features that will later be used as a features i.e. variables for classification algorithms.

Various transforms like Wavelet, Hilbert \cite{madeiro2012innovative}, Empirical Mode Decomposition \cite{pal2012empirical} and similar are proposed. These transformations are performed to transform the original signal from time-domain to the frequency-domain. Frequency characteristics of the particular signal sections are then used to build upon the decision logic that can identify particular QRS segment i.e. heartbeats.
Results reported in recent works show very high detection rates in sensitivity and positive predictivity, both more than 99.5\% \cite{zidelmal2012qrs}.

This kind of transformations alter original ECG morphology. E.g. Hilbert transform can be applied in the frequency domain to simply shifts all positive frequency components by -90\degree and all negative frequency components by +90\degree. The amplitude always remains constant throughout this transformation. Aplying the first derivative on the signal followed by the Hilbert transform does not detect QRS complex, but it only aims to enhance the peaks and in that way ensure bigger probability to find the QRS region by threshold-detection rules \cite{rezk2011algebraic}.

Since the ECG signal is actually a time series, conventional statistical and data mining procedures are not directly applicable due to the violation of independence between the observations. This dependence or to be more precise, existence of particular state transition probabilities between the observations makes the ECG analysis problem suitable for application of Hidden Markov Models (HMM) statistical method. In the HMM process, the result of the previous state will influence the result of the following state. This is similar to the processes in the heart which posses properties of successive stage transitions. This makes HMM models a promising approach for ECG analysis and notable results were achieved in QRS detection with some advancements in pathology classification \cite{HMM_6409718}.

\subsection{HRV Analysis}

Heart rate variability analysis (HRV) implies advanced analytical models based on chaos theory,  statistics, entropy and other features that describe changes in the heart rate \cite{Jovic_ITI_2009_5196051}. Main motivation for this approach is the fact that pathological states of the heart muscle have impacts on the heart rhythm. As it is reported in the literature, healthy heart has in fact more chaotic behaviour that the pathologically affected heart \cite{bolis1999autonomic}. As this approach is based on the heart rhythm analysis, QRS detection is assumed and for the development and testing purposes, in this kind of research annotated databases are used to calculate linear and non-linear rhythm features. Those features are then used as inputs for different machine learning algorithms i.e. classifiers. Recent work shows significant advancements and potential of HRV in diagnosis of the various pathologies and arrhythmias which implies that reliable and precise QRS detection is mandatory for this kind of analysis \cite{jovic2011electrocardiogram}.

\subsection{Testing Sources}

Standardization in the field of computer aided ECG analysis originates from 1980. under the patronage of the European Commission with the project called „Common Standards for Quantitative Electrocardiography (CSE)“ \cite{willems1990common}. Algorithms that are presented to the scientific community are evaluated with parameters that represent their ability to successfully identify QRS complexes and this is covered by ANSI \cite{american2002cardiac}. To enable comparison of the tested algorithms, common databases are used like e.g. MIT-BIH Arrhytmia Database and a web site where those can be found is called Physionet \cite{goldberger2000physiobank}. ECG records in those databases are annotated which means that there is information about the occurrences of various ECG artefacts like normal heart beat, PVC (Premature Venticular Contraction), changes in signal quality etc. These databases include many beats and many pathological states, various noise occurrences and in that way represent a good testing ground.

\subsection{Performance Measures}

Although most of the papers indicate algorithm results based on classical measures like sensitivity (Se), specificity (Sp), positive predictivity (+P), detection error rate (DER), area under the Receiver Operating Characteristics (ROC) curve (AUC) and overall accuracy (ACC) \cite{manikandan2012novel} there are still some doubts on measures that are really good indicators for algorithms' performance. All of these metrics rely on the calculations that show how good a classifier is in detecting true positives (TP - class detected is actually the real class) and true negatives (TN - e.g. QRS complex is not reported when there is no one in the testing signal) and at the same time prone to mistakes of false positives (FP) and false negatives (FN) erroneous classifications. Researchers need to be careful in reporting the performance measures to ensure that results are not influenced by the sampling rate or \emph{a priori} probability of considered class \cite{Darrington2009}. Performance measures are defined as

\begin{equation}
Se(\%) = 1- \dfrac{FN}{TP+FN}=\dfrac{TP}{TP+FN} * 100\%
\end{equation}

\begin{equation}
Sp(\%) = \dfrac{TN}{N}=\dfrac{TN}{TN+FP} * 100\%
\end{equation}

\begin{equation}
+P(\%) = 1- \dfrac{FP}{TP+FP}=\dfrac{TP}{TP+FP} * 100\%
\end{equation}

\begin{equation}
DER(\%) = \dfrac{FP+FN}{TP+FN} * 100\%
\end{equation}

\begin{equation}
ACC(\%) = \dfrac{TP}{TP+FP+FN} * 100\%
\end{equation}

In spite of the apparently clear testing methodology there is still a problem with reporting results and comparison on the different algorithms performance. Some authors test their algorithms on some of the records and not on the whole databases and often authors exclude some parts of the signal e.g. ventricular flutter episodes or parts with high noise like in the record 207 from the MIT-BIH Arrhythmia Database \cite{manikandan2012novel}.

\subsection{Recent Approaches}

Recent approaches to more robust and more realistic tests include class-based and subject-based testing. In class-based testing all records are used for training of the classifier. Drawback of this approach is that it is not realistic in a way that, when it comes to a prediction, waveforms that should be predicted are from a patient (record) that was used for learning of the classifier. Newer and more realistic evaluation method proposed in only few papers is subject-based evaluation. In subject-based testing, whole records are excluded from the training set and predictions for particular record are made in respect to that i.e. based on the other records from the database. This situation simulates real world scenario when software application that implements a classifier faces new patient it has never "seen" before. Results in subject-based testing are always lower than in class based testing \cite{ye2012heartbeat}.
Problem with these approaches is that often researches use limited number of classes and variety of the ECG morphology shapes in real life is much higher. However, ANSI/AAMI defined five-class division so evaluations done in that way are actually following a standard \cite{american2002cardiac}.

\subsection{Bullseye Testing}

Since ECG signal can have numerous varieties, here we suggest new kind of test learned from somewhat similar discipline of shape recognition - the bullseye test. In shape recognition, bullseye testing can be used to evaluate how good a classifier is in finding similar shapes from the testing database. Bullseye test can be used for supervised and for unsupervised learning procedures. 

In supervised learning, when testing database includes annotation or a class for each shape, every shape is matched with all shapes in the database. For example, if given dataset includes 20 instances of each class, we use proposed matching algorithm to identify e.g. top 40 matches (double the number of class instances) and discard the others. Among these 40 we count correct matches for each class in question. The accuracy of shape retrieval is the ratio of the number of correct hits to the highest possible correct hits \cite{donoser2010efficient}, \cite{lin2008efficient}.

In unsupervised learning, e.g. when database is very big and not all classes are identified in the training or testing dataset, bullseye test can be performed to find the most similar shapes and then visual inspection can be made and results reported as a graphic \cite{kontschieder2010beyond}.

\begin{figure}[h]
	\centerline{\psfig{figure=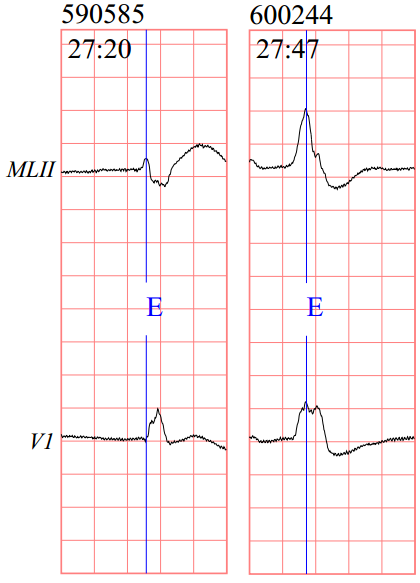,width=50mm} }
	\caption{Example of the ECG waves from record 207, MIT-BIH AD database where different wave morphologies share same annotation}
	\label{fig:sample_graph}
\end{figure}

Since similar problem exists in many ECG databases, i.e. not all ECG segments are annotated or annotations are not detailed enough, bullseye test could be performed and results visually reported. In that way one could see how various morphologies are matched e.g. PVCs of the similar shapes and orientation should be grouped together. If the algorithm that is tested groups visually similar PVC beats together, than that could mean that the algorithm is really distinguishing morphologies correctly and that decision about the class in question is not made on only morphology invariant features like entropy, duration, frequency components etc. Figure 2 shows an excerpt of the MIT-BIH Arrhythmia Database where one annotation implies in fact two different morphologies.
Results of a proposed bullseye test of the ECG waves morphology classification is shown on the figure 3.

\begin{figure}[h]
	\centerline{\psfig{figure=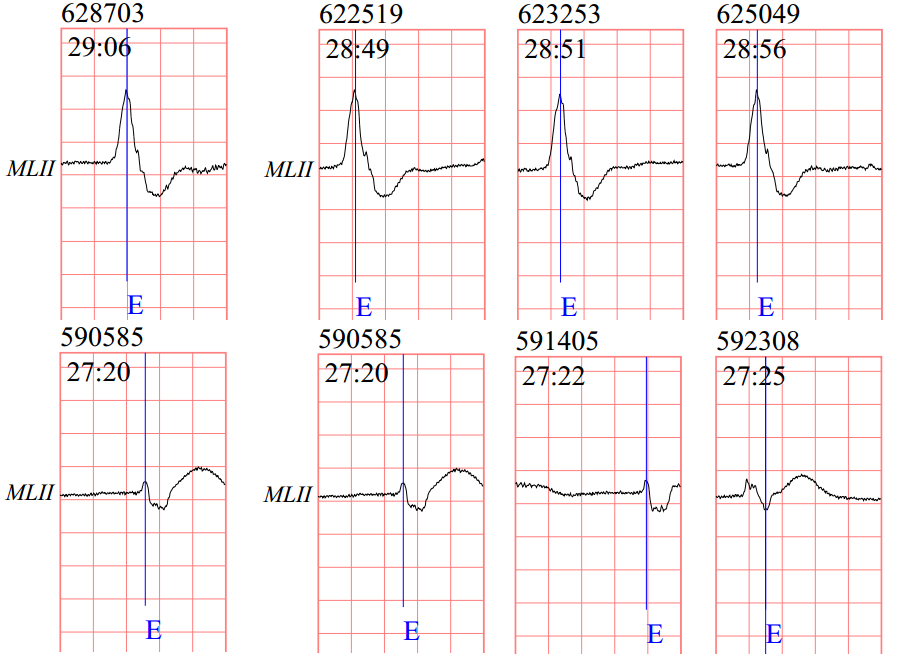,width=68.7mm} }
	\caption{Exemplar results of the proposed bullseye test. Upper row shows the best three matches for one morphology and the second row for other morphology of the same annotation}
	\label{fig:sample_graph}
\end{figure}

\section{Challenges and Gaps}

Despite very advanced and fast heartbeat detectors and even recent discoveries in ECG morphology analysis, there is limited use of those algorithms in clinical practice. To gain understanding in why is it so, besides all possible business related issues and medical clearances that must be satisfied, here we will try to identify medicine related issues.

\subsection{Morphology Considerations}

To understand what kind of algorithms are needed for efficient clinical practice in ECG analysis and diagnosis, one must first understand what is considered to be a normal heart rhythm. Normal heart rhythm i.e. behaviour, considering medical literature, must satisfy all of the following four criteria \cite{Najeeb}:

1)  Heart Rate: Heart rate should be between 60 and 100 beats per minute \cite{de2008basic}, \cite{taylor2008150},  \cite{Gacek2012}.

2)  Origin: Origin of the particular beat i.e. electrical impulse must be in the SA node \cite{Clifford:2006}, \cite{taylor2008150}, \cite{Gacek2012}.

3)  Pathway: Impulse must propagate throughout the normal conducting pathway \cite{Clifford:2006}.

4)  Speed: Impulse must propagate at the normal speed (i.e. speeds) \cite{Clifford:2006}.

Here we can see that heart rate is just one of the four criteria for identifying the normal heart behaviour. Of course, as it is explained earlier, significant advancements are achieved in identifying heart work problems from heart rate variability analysis, but other three criteria have more impact on the ECG wave morphology than on the heart rate itself. Considering the underlying ECG mechanism explained earlier, we can realize that if the second criteria is not met than the QRS complex will not proceed after exactly one P wave or P waves resulting from other myocardial cells, e.g. ectopic myocardial cell, will not be normally periodic or maybe P wave could be reversed due to the abnormal impulse propagation direction [52]. If the electric impulse is not propagating throughout the heart in normal conducting pathway than ECG signal can show short P-R segment due to the lack of the AV node pause or various morphology deviations can be symptoms of the abnormal ventricle contraction due to the bypass or skip of the fast Purkinje system and propagation throughout the much slower myocardial cells \cite{Clifford:2006}. This is example how abnormal conducting pathway can in turn lead to the abnormal speed of the signal propagation. Often in heart work problems, one issue can cause cascading problems resulting from collateral damages that just multiply abnormal behavior with each following beat. This is why it is very important to make correct and early diagnosis of the problem.

\subsection{P Wave}

P wave detection and classification is a problem because of the wave's small amplitude and attenuation due to the filtering of the signal. However, P wave is an important component of the clinical ECG diagnosis process since it can indicate various atrial problems and not just by its frequency of occurrence \cite{hampton2003150}, [53] but also by its shape \cite{de2008basic}, \cite{hampton2003150}. Besides frequency and morphology, P wave direction can also indicate pathological states like dextrocardia \cite{hampton2003150}.

\subsection{T Wave}

T wave morphology changes are also very important in pathologic states diagnosis. Different shape of the T wave can indicate problem with beat origin and re-polarization issue due to the branch block. Normal or abnormal T wave shape depending on the visibility of P wave and narrowness of the QRS can indicate heart block problems and inverted T wave can indicate among other problems e.g. right ventricular hypertrophy \cite{hampton2003150}.

\subsection{Multi-Lead Analysis}

Based on the morphological features of the ECG and keeping in mind the underlying electro-mechanics in 3D space, experienced cardiologist can identify not just the problem in the heart muscle but also the approximate region of the heart that is influenced by the problem e.g. ischemic region and even locate where is the source of the problem e.g. approximate source of the ectopic beat. In that way, ECG can be used as a low cost and fast tool for beginning of the diagnosis process.
Multi-lead analysis is crucial for a wide range of pathology identifications \cite{hampton2003150} and serious work in this direction is just appearing in the literature \cite{ye2012heartbeat}, \cite{Chang20123165}.

\subsection{ST Segment}

Acute coronary syndrome (ACS) is a significant health problem in industrialized countries and is becoming an increasingly significant problem in developing countries. ACS is a clinical syndrome defined by characteristic symptoms of myocardial ischemia in association with ECG ST-segment morphology changes (elevation or depression) indicative of the occlusion of a major epicardial coronary artery \cite{Kushner2013178}.

\subsection{Individual Adjustments}

Another challenge in computer aided ECG analysis is a fact that ECG is considered biometric characteristic which means that every person has individual ECG signature. Although most of the people have similar ECG manifestation there is part of population which has significant deviations in normal ECG or in pathological states \cite{douglas2006temporal}. Algorithms that could cover those cases should be capable of learning what is in fact normal wave morphology for each particular patient and then report identified misalignments if they occur.

\subsection{Annotations and Testing}

Problem with comparison of different approaches and algorithms in ECG analysis derives from the fact that algorithms' capabilities evolved over the QRS detection problem and even more powerful and precise algorithms are expected to arise in order to address the above mentioned challenges. Although group of researchers gathered around the Physionet made a lot of software components available for testing of the newly created algorithms \cite{goldberger2000physiobank}, recent work identifies gaps between annotation standards \cite{Clifford:2006}. Considering various upcoming challenges described earlier in this paper, new meta-model of annotations and e.g. online testing components that could unify various databases are welcome. Furthermore, authors of this paper think that new testing standards should be developed to meet the upcoming challenges so that researchers and their results can be more close to the real clinical use and benefits for patients which should be the ultimate goal of the work in this area.

\section{Future Work}

Based on the state of the art literature review and medical literature inspection, we can conclude that future work in this field should be focused on the efforts to further standardize annotations and testing principles and methodologies. Considering algorithmic efforts, further advancements should be done towards the algorithms that could correctly classify ECG wave morphologies and towards multi-lead analysis and decision modules to get more near the clinical use of the proposed systems. Supervised learning algorithms that can correlate particular wave morphology to a pathological state are welcome as are the unsupervised learning algorithms that could be adaptable to individual patients and detect outlying waves even if the particular morphology is not associated with the pathology in the knowledge base. Considering the latest advancements in HRV analysis which is based on QRS detection, one can assume what would be potential benefits and possible research opportunities if more powerful morphology classification algorithms will be developed. In that way, researches could utilize their findings from chaos theory and non-linear systems to conduct ECG morphology variability (EMV) analysis. All these findings could lead to the reliable mobile and tele-medical solutions whose prototypes are already developed but with very limited analysis capabilities \cite{vrcek2007integrated}, \cite{5503907}, \cite{Batistatos20121140}, \cite{Naikc_5546468}.

\section{Conclusion}

Despite growing technology in cardiology the electrocardiogram (ECG) stays as a cheap and quickly tool for beginning the identification of potential lethal cardiac pathology.  The ECG analysis except the recognition of basic waves and complexes (P, QRS and T, sometimes U) keep in mind also its duration, configuration and orientation (positive or negative) and formation of parts between complexes and waves (for example ST segment which is crucial in identification of acute myocardial infarction).  The complexity of ECG curve embarrasses computer programmers in finding the best mathematical model to describe what is exactly happening in cardiac cycle and that is the reason why we still do not have a software solution for ECG analysis which is comparable to clinical decision. Any new model in describing ECG curve is invaluable for future development of better software programs for ECG analysis which is used not only in ECG machine than in many cardiology devices (for example monitors and pacemakers) and for better understanding the ECG in various clinical conditions. Also, new testing methods and records annotated in more detail that will follow the algorithmic advancements are welcome.

\bibliographystyle{IEEEtran}

\bibliography{IEEEabrv,IEEEexample}

\end{document}